\documentclass[11pt,a4paper]{article}
\usepackage[hyperref]{eacl2021}
\usepackage{times}
\usepackage{latexsym}

\usepackage{microtype}

\aclfinalcopy % Uncomment this line for the final submission

\usepackage{soul}
\usepackage{float}
\usepackage{multirow}
\usepackage{multibib}
\usepackage{booktabs}
\usepackage{graphicx}
\usepackage[noend]{algorithm2e}
\usepackage{tabularx}
\usepackage{fontawesome5}
\usepackage[mode=buildnew]{standalone}
\usepackage{tikz}
\usepackage[tbtags]{amsmath}

\usepackage{algorithm2e}
\usepackage{xcolor}

\newcommand{\xvar}[1]{\text{#1}}

\newcommand{\xvbox}[2]{\makebox[#1][l]{#2}}

\SetAlFnt{\footnotesize}

\SetAlCapNameFnt{\footnotesize}
\SetAlCapFnt{\scriptsize}

\SetAlCapSty{xAlCapSty}

\SetCommentSty{xCommentSty}
\LinesNumbered
\SetSideCommentRight
\DontPrintSemicolon
\RestyleAlgo{algoruled}

\DeclareMathOperator*{\argmax}{arg\,max}

\title{Adversarial Stylometry in the Wild: \\ Transferable Lexical Substitution Attacks on Author Profiling}

\author{Chris Emmery \\
  CSAI, Tilburg University \\ CLiPS, University of Antwerp \\
  \texttt{cmry@pm.me} \\\And
  \'{A}kos K\'{a}d\'{a}r \\
  Borealis AI \\
  \texttt{akos.kadar@}\\ \texttt{borealisai.com}\\\And
  Grzegorz Chrupa\l{}a \\
  CSAI, Tilburg University \\
  \texttt{\ g.a.chrupala@uvt.nl}}

\date{}
\begin{document}
\maketitle

\begin{abstract}

Written language contains stylistic cues that can be exploited to automatically infer a variety of potentially sensitive author information. Adversarial stylometry intends to attack such models by rewriting an author's text. Our research proposes several components to facilitate deployment of these adversarial attacks in the wild, where neither data nor target models are accessible. We introduce a transformer-based extension of a lexical replacement attack, and show it achieves high transferability when trained on a weakly labeled corpus---decreasing target model performance below chance. While not completely inconspicuous, our more successful attacks also prove notably less detectable by humans. Our framework therefore provides a promising direction for future privacy-preserving adversarial attacks.

\end{abstract}

\section{Introduction}

The widespread use of machine learning on consumer devices and its application to their data has sparked investigation of security and privacy researchers alike in correctly handling sensitive information \cite{edwards2015censoring,abadi2016deep}. Natural Language Processing (NLP) is no exception \cite{fernandes2019generalised,li2018towards}; written text can contain a plethora of author information---either consciously shared or inferable through stylometric analysis \cite{rao2000can,adams2006classification}. This characteristic is fundamental to author profiling \cite{koppel2002}, and while the field's main interest pertains to the study of sociolinguistic and stylometric features that underpin our language use \cite{daelemans2013explanation}, herein simultaneously lie its dual-use problems. Author profiling can, often with high accuracy, infer an extensive set of (sensitive) personal information, such as age, gender, education, socio-economic status, and mental health issues \cite{eisenstein2011,alowibdi2013,volkova2014,plank2015,volkova2016}. It therefore potentially exposes anyone sharing written online content to unauthorized information collection through their writing style. This can prove particularly harmful to individuals in a vulnerable position regarding e.g., race, political affiliation, or mental health.

Privacy-preserving defenses against such inferences can be found in the field of adversarial\footnote{These are adversarial attacks on models making stylometric predictions, not to be confused with adversarial learning.} stylometry. Our research\footnote{All code, data, and materials to fully reproduce the experiments are openly available at \url{https://github.com/cmry/reap}.} concerns the obfuscation subtask, where the aim is to rewrite an input text such that the style changes, and stylometric predictions fail. It is part of a growing body of research into adversarial attacks on NLP \cite{smith2012adversarial}, which various modern models have proven vulnerable to; e.g., in neural machine translation \cite{ebrahimi2018adversarial}, summarization \cite{cheng2018seq2sick}, and text classification \cite{liang2018deep}.

Adversarial attacks on NLP are predominantly aimed at demonstrating vulnerabilities in existing algorithms or models, such that they might be fixed, or explicitly improved through adversarial training. Consequently, most related work focuses on white or black-box settings, where all or part of the target model is accessible (e.g., its predictions, data, parameters, gradients, or probability distribution) to fit an attack. The current research, however, does not intend to improve the targeted models; rather, we want to provide the attacks as tools to protect online privacy. This introduces several constraints over other NLP-based adversarial attacks, as it calls for a realistic, in-the-wild scenario of application. 

Firstly, authors seeking to protect themselves from stylometric analysis cannot be assumed to be knowledgeable about the target architecture, nor to have access to suitable training data (as the target could have been trained on any domain). Hence, we cannot optimally tailor attacks to the target, and need an accessible method of mimicking it to evaluate the obfuscation success. To facilitate this, we use a so-called substitute model, which for our purposes is an author profiling classifier trained in isolation (with its own data and architecture) that informs our attacks. Attacks fitted on substitute models have been shown to transfer their success when targeting models with different architectures, or trained on other data, in a variety of machine learning tasks \cite{papernot2016transferability}. The effectiveness of an attack fitted on a substitute model when targeting a `real' model is then referred to as \emph{transferability}, which we will measure for the obfuscation methods proposed in the current research. 

Secondly, for an obfuscation attack to work in practice (e.g., given a limited post history), it should suggest relevant changes --to-- the author's writing \emph{on a domain of their choice}. This implies the substitute models should be fitted locally, and therefore need to meet two criteria: reliable access to labeled data, and being relatively fast and easy to train. To meet the first criterion, the current research focuses on gender prediction, as: i) Twitter corpora annotated with this variable are by far the largest (and most common), ii) author profiling methods typically use similar architectures for different attributes; therefore, the generalization of attacks to other author attributes can be assumed to a large extent, and, most importantly, iii) \newcite{beller2014ma} and \newcite{emmery2017simple} have shown that through distant labeling, a representative corpus for this task can be collected in under a day. This allows us to measure transferability of attacks fitted using realistically collected distant corpora to models using high-quality hand labeled corpora.

\begin{figure}
\includegraphics[width=\columnwidth, clip,]{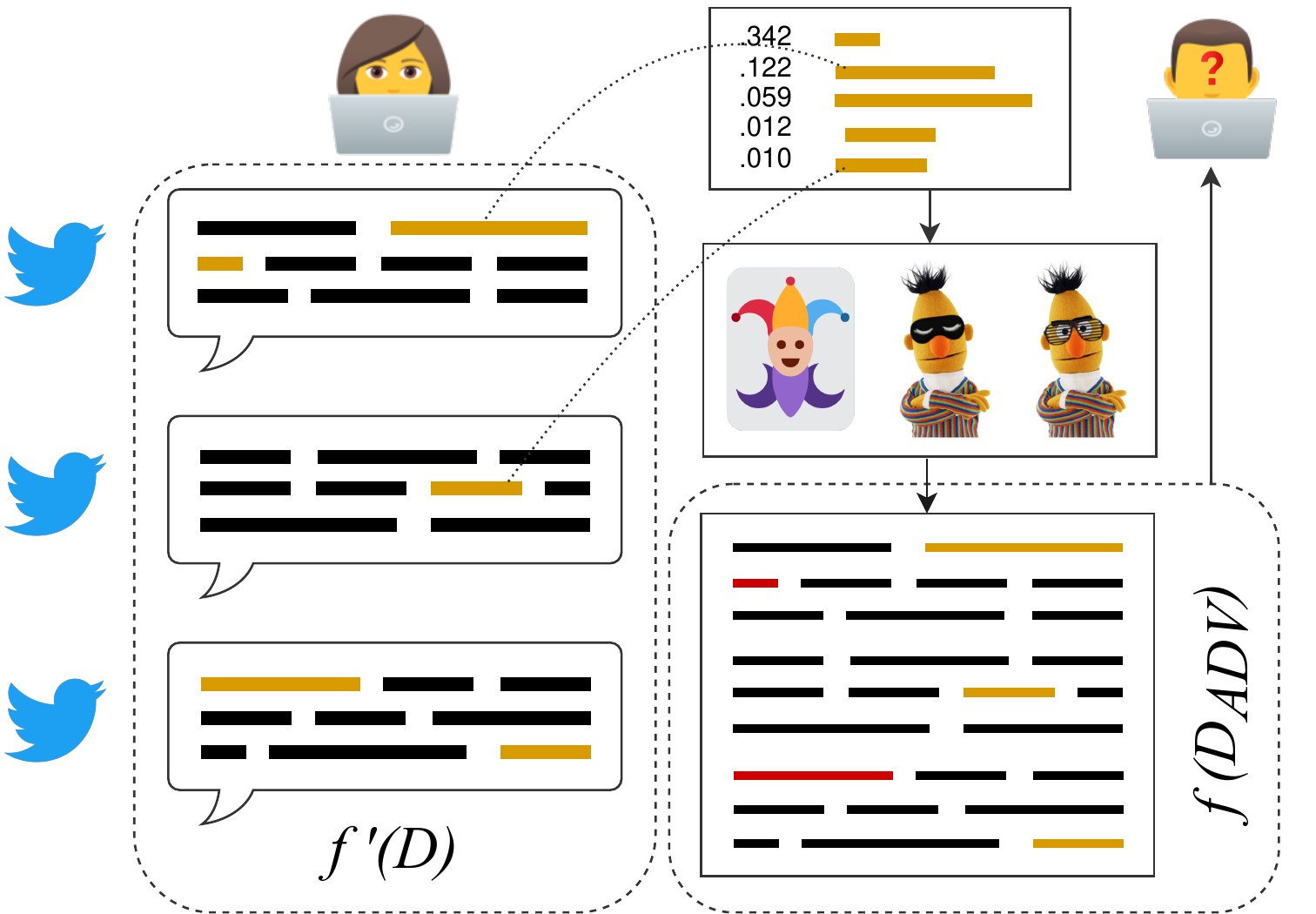}
\caption{Obfuscation scenario: model $f'$ trains on tweet batches, an omission score is used to determine and rank the words according to their classification contribution. These are then passed to either TextFooler, Masked BERT, or Dropout BERT to suggest top-$k$ replacement candidates. From these, a selection is made based on their class probability change on $f'(D)$. Finally, $f$ is evaluated on the perturbed tweets $D_{\textsc{adv}}$.}
\label{fig:obf}
\end{figure}

As for the attacks, we focus on lexical substitution of content words strongly related to a given label, as those have been shown to explain a significant portion of the accuracy of stylometric models \cite[see e.g.,][]{rao2000can,burger2011,sap2014,rangel2016}. To that effect, we extend the substitution attack of \newcite{jin2019bert} and apply it to author attribute obfuscation. Specifically, we explore the potential of training a simple (as to meet the speed criterion), non-neural substitute model $f'$ to indicate relevant words to perturb, where retaining the original meaning is prioritized. Two transformer-based models are introduced to the framework to propose and rank lexical substitutions towards a change in the predictions of $f'$. We evaluate if the attacks on $f'$ transfer across corpora, architectures, and a separately trained target model $f$ (see Figure~\ref{fig:obf}). Finally, we measure the quality of changes using automatic evaluation metrics, and conduct an human evaluation that focuses on detection accuracy of the attacks.

\section{Related Work}

Stylometry, the study of (predominantly) writing style, dates back  several decades \cite{mosteller1963inference}, and has seen increased accessibility through the introduction of statistical models \cite[see surveys by ][]{holmes1998evolution,neal2017surveying} and machine learning \cite[e.g.,][]{matthews1993neural,merriam1994neural}. Computational stylometry distinguishes several subtasks such as determining \cite[][]{baayen2002experiment} and verifying author identity \cite[][]{koppel2004authorship}, and author profiling \cite{argamon2005lexical}; e.g., predicting demographic attributes. Adversarial stylometry (as conceptualized by \citeauthor{brennan2012adversarial}, \citeyear{brennan2012adversarial}) intends to subvert these inferences by changing an author's text through imitation, or, as pertains to our research, the obfuscation of writing style \cite{kacmarcik2006obfuscating,caliskan2015coding,le2015secure,xu2019alter}.

These changes, or perturbations, can be produced in several ways, and the task is therefore often conflated with paraphrasing \cite{reddy2016obfuscating}, style transfer \cite{kabbara2016stylistic}, and generating adversarial samples or triggers \cite{zhang2020adversarial}. Regardless of the employed method, the main challenge of obfuscation lies in retaining the original meaning of an input text; its written language medium limits any perturbations to discrete outputs, and unnatural discrepancies are significantly better discernible by humans than, say, a few pixel changes in an image. An additional, persistent limitation is the absence of evaluation metrics that guarantee complete preservation of the original meaning of the input whilst changes remain unnoticed \cite{potthast2016}. This not only inhibits automatic evaluation of obfuscation, but all natural language generation research \cite{novikova2017we}---placing an emphasis on human evaluation \cite{van2019best}.

It is perhaps for this reason that most obfuscation work uses heuristically-driven, controlled changes such as splitting or merging words or sentences, removing stop words, changing spelling, punctuation, or casing \cite[see e.g.,][]{karadzhov2017case,eger2019text}. These specific attacks are typically easier to mitigate through preprocessing \cite{juola2011analyzing}.  Obfuscation through lexical substitution \cite{mansoorizadeh2016author,bevendorff2019heuristic,bevendorff2020divergence} provides a middle ground of control, semantic preservation and attack effectiveness; however, they might prove less effective against models relying on deeper stylistic features (e.g.\ word order, part-of-speech (POS) tags, or reading complexity scores). End-to-end systems have been employed for similar purposes \cite{shetty2018a4nt,saedi2020}, or to rewrite entire phrases \cite{emmery2018style,bo2019er} using (adversarially-driven) autoencoders. Such attacks seem less common, and provide less control over the perturbations and semantic consistency.

Our work does not assume the attacks to run end-to-end, but with a hypothetical human in the loop. We further opt for techniques that are more likely to find strong semantic mirrors to the original text while making minimal changes. A substitute model (the algorithm, hyper-parameters, and output of which an author can manipulate as desired) is employed to indicate candidate replacement words, and our attacks suggest and rank those against this substitute. Moreover, prior work typically attacks adversaries trained on the same data, whereas we add a transferability measure. Lastly, while author identification has been investigated in the wild \cite{stolerman2013classify}, our work is, to our knowledge, the first to make a conscious effort towards realistic applicability of obfuscation techniques.

\section{Method}

Our attack framework extends TextFooler \cite[TF,][]{jin2019bert} in several ways. First, a substitute gender classifier is trained, from which the logit output given a document is used to rank words by their prediction importance through an omission score (Section~\ref{subs:twi}). For the top most important words, substitute candidates are proposed, for which we add two additional techniques (Section~\ref{subs:att}). These candidates can be checked and filtered on consistency with the original words (by their POS tags, for example), accepted as-is, or re-ranked (Section \ref{subs:checks}). For the latter, we add a scoring method. Finally, the remaining candidates are used for iterative substitution until TF's stopping criterion is met (i.e., the prediction changes, or candidates run out).

\subsection{Target Word Importance} \label{subs:twi}

We are given a target classifier $f$, substitute classifier $f'$, a document $D$ consisting of tokens $D_i$, and a target label $y$. Here, $f'$ is trained on some corpus $X$, and receives an author's new input text $D$, where the author provides label $y$. We denote a class label as $\bar{y}$ if $f'(D)$ predicts anything but $y$. Our perturbations form adversarial input $D_\textsc{adv}$, that intends to produce $f'(D_\textsc{adv}) = \bar y$, and thereby implicitly $f(D_\textsc{adv}) = \bar y$. Note that we only submit $D$ to $f$ for evaluating the attack effectiveness, and it is never used to fit the attack itself.

To create $D_\textsc{adv}$, a minimum number of edits is preferred, and thus we rank all words in $D$ by their omission score \cite[similar to e.g.,][]{kadar2017representation} according to $f'$ (\texttt{omission\_score} in Algorithm~\ref{alg:obf}). Let $D_{\setminus i}$ denote the document after deleting $D_i$, and $o_y(D)$ the logit score by $f'$. The omission score is then given by $o_y(D) - o_y(D_{\setminus i})$, and used in an importance score $I$ of token $D_i$, as:

\begin{equation}
\label{eq:words_importance}
    I_{D_i}=
    \begin{cases}
      o_y(D) - o_y(D_{\setminus i}), \\ \quad \ \  \ \text{if}\ f'(D)=f'(D_{\setminus i}) = y. \\
      o_y(D) - o_y(D_{\setminus i}) + o_{\bar{y}}(D) - o_{\bar{y}}(D_{\setminus i}), \\ \quad \ \ \   
      \text{if}\ f'(D) = y, f'(D_{\setminus i})= \bar y, y\neq \bar y.
    \end{cases}
\end{equation}

\noindent With $I_{D_i}$ calculated for all words in $D$, the top $k$ ranked tokens are chosen as target words $T$.

\begin{algorithm}

\SetKwFunction{cumprod}{cumprod}
\SetKwFunction{length}{length}
\SetKwFunction{zeros}{zeros}
\SetKwFunction{ceil}{ceil}

\SetKwInOut{Input}{Input}
\SetKwInOut{Output}{Output}

\caption{Obfuscation by lexical replacement.\label{alg:obf}}

\Input{%
		\hspace{2mm} \xvbox{2mm}{$f'$} -- substitute model \\
		\hspace{2mm} \xvbox{31mm}{$D = \{w_0, w_1, \ldots, w_n\}$} -- document \\
		\hspace{2mm} \xvbox{2mm}{$y$} -- target label \\
		\hspace{2mm} \xvbox{9mm}{\xvar{checks}} -- apply checks (bool) \\
		\hspace{2mm} \xvbox{2mm}{\xvar{k}} -- target max $k$-amount words \\
	  }
\Output{
		\hspace{2mm} \xvbox{7mm}{$D_\textsc{adv}$} -- obfuscated document
	   }

  \BlankLine
          
  \For{$D_i \in D$ }{
    \tcp{via Equation~\ref{eq:words_importance}}
    $I_{D_i} \leftarrow$ \texttt{omission\_score}($f'$, $y$)
	
  }
  \BlankLine
  $T \leftarrow$ \texttt{top\_k}(\texttt{argsort\_desc}($D, I_{D_i}$ scores), k)
  \BlankLine
  $D_\textsc{adv} = D$ \\
  \For{$t \in T$ }{
    \tcp{substitution attack on $t$}
    $C_t \leftarrow$\texttt{candidates}($t$) \\
    $A = ( D_{\textsc{adv}\ {1:i-1}}, C_{t, j}, D_{\textsc{adv}\ {i+1:n}} )_{1 \leq j \leq |C_t|}$ \\
    $\bar{A} =$ \texttt{filter/rank}($D$, $A;t$, checks) \\
    \tcp{test attack success on $f'$}
    \For{$D' \in \bar{A}$ }{
        \uIf{$\argmax o_y(D') \neq y$}{\Return $D_\textsc{adv} = D'$}
        \uElseIf{$o_y(D') < o_y(D_\textsc{adv})$}{
                $t$ in $D_\textsc{adv}$ is replaced with $c$ from $D'$  
        }
    }
}
\Return $D_\textsc{adv}$
\end{algorithm}

\subsection{Lexical Substitution Attacks} \label{subs:att}

Four approaches to perturb a target word $t \in T$ are considered in our experiments. These operations are referred to as \texttt{candidates} in Algorithm~\ref{alg:obf}.

\paragraph{Synonym Substitution (WS)}

This TF-based substitution embeds $t$ as $\boldsymbol{t}$ using a pre-trained embedding matrix $\boldsymbol{V}$. $C_t$ is selected by computing the cosine similarity between $\boldsymbol{t}$ and all available word-embeddings $\boldsymbol{w} \in \boldsymbol{V}$. We denote cosine similarity with $\Lambda(\boldsymbol{t}, \boldsymbol{w})$. A threshold $\delta$ is used to keep only reliable candidates $\Lambda(\boldsymbol{t}, \boldsymbol{w}) > \delta$.

\paragraph{Masked Substitution (MB)}

The embedding-based substitutions can be replaced by a language model predicting the contextually most likely token. BERT \cite{devlin2018bert}---a bi-directional encoder \cite{vaswani2017attention} trained through masked language modeling and next-sentence prediction---makes this fairly trivial. By replacing $t$ with a mask, BERT produces a top-$k$ most likely $C_t$ for that position. Implementing this in TF does imply each previous substitution of $t$ might be included in the context of the current one. This method of contextual replacement has two drawbacks: i) semantic consistency with the \emph{original} word is not guaranteed (as the model has no knowledge of $t$), and ii) the replaced context means semantic drift can occur, as all subsequent substitutions follow the new, possibly incorrect context.

\paragraph{Dropout Substitution (DB)} A method to circumvent the former (i.e., BERT's masked prediction limitations for lexical substitution), was presented by \newcite{zhou2019bert}. They apply dropout \cite{srivastava2014dropout} to BERT's internal embedding of target word $t$ before it is passed to the transformer---zeroing part of the weights with some probability. The assumption is that $C_t$ (BERT's top-$k$) will contain candidates closer to the original $t$ than the masked suggestions.

\paragraph{Heuristic Substitution}

To evaluate the relative performance of the techniques we described before, we employ several heuristic attacks as baselines. In the order of Table~\ref{tab:results}: {\underline{1337}-speak}: converts characters to their leetspeak variants, in a similar vein to e.g.\ diacritic conversion \cite{belinkov2018synthetic}. {Character \underline{flip}}: inverts two characters in the middle of a word, which was shown to least affect readability \cite{rayner2006raeding}. {Random \underline{space}s}: splits a token into two at a random position.

\subsection{Candidate Filtering and Re-ranking} \label{subs:checks}

Given $C_t$, either all, or only the highest ranked candidate can be accepted as-is. Alternatively, all $D'$ can be filtered by submitting them to checks, or re-ranked based on their semantic consistency with $D$. These operations are referred to as \texttt{rank/filter} in Algorithm~\ref{alg:obf}---both of which can be executed.

\paragraph{Part-of-Speech and Document Encoding}

TF employs two checking components: first, it removes any $c$ that has a different POS tag than $t$. If multiple $D'$ exist so that $f'(D') = \bar y$, it selects the document $D'$ which has the highest cosine similarity to the Universal Sentence Encoder (USE) embedding \cite{cer2018universal} of the original document $D$. If not, the $D'$ with the \emph{lowest} target word omission score is chosen (as per TF's method).

\begin{table*}
    \centering
    \resizebox{\textwidth}{!}{
    \begin{tabular}{@{\extracolsep{\fill}}l|rr|rr|rr|rr|r}
    \toprule
                                       & \textsc{authors}   & \textsc{tweets}        & \textsc{female}    & \textsc{male}      & \textsc{train}     & \textsc{test}      & \textsc{tokens}        & \textsc{types}  & \textsc{avg size}  \\
    \midrule
    \citeauthor{huang2020-lrec}        & 37,929  & 47,211        & 26,758    & 20,453    & 30,602    &  7,651    & 935,062       & 46,600     & 28  \\
    \citeauthor{emmery2017simple}      & 6,610   & 16,788,612    & 61,736    & 32,900    & 75,918    & 18,718    & 146,736,657   & 9,942,399  & 301 \\
    \citeauthor{volkova2015inferring}  & 4,620   & 12,226,859    & 32,376    & 26,708    & 47,298    & 11,777    & 67,186,535    & 7,836,539  & 269 \\
    \bottomrule
    \end{tabular}}
    \caption{Corpus statistics indicating the number of authors, tweets, female and male labels, the size of the train and test splits, number of types (unique words) and tokens (total words), and average tokens per document (avg size). }
    \label{tab:data}
\end{table*}

\paragraph{BERT Similarity} \label{p:bertsim}

\newcite{zhou2019bert} use the concatenation of the last four layers in BERT as a sentence's contextualized representation $\boldsymbol{h}$. We apply this in both Masked (MB) and Dropout (DB) BERT to re-rank all possible $D'$ by embedding them. Given document $D$, target $t$, and perturbation candidate document $D'$, $C_t$ would be ranked via an embedding similarity score:
\begin{equation}
\begin{split}
& \textsc{sim}\left(D, D^{\prime} ; t\right) =  \\ \sum_{i}^{n} w_{i, t} \times
& \Lambda\left(\boldsymbol{h}\left(D_{i} | D \right), \boldsymbol{h}(D_{i}^{\prime} | D^{\prime}) \right)
\end{split}
\end{equation}
where $\boldsymbol{h}\left(D_{i} | D\right)$ is BERT's contextualized representation of the $i^{th}$ token in $D$, and $w_{i, t}$
is the average self-attention score of all heads in all layers ranging from the $i^{th}$ token with respect to $t$ in $D$.\footnote{Zhou et al. (2019) additionally use a proposal score for finding $T$ that we replaced with the omission score.}

\section{Experiment}

\subsection{Data} \label{subs:data}

We use three author profiling sets (see Table~\ref{tab:data} for statistics) that are annotated for binary gender classification (male or female): first, that of \newcite{volkova2015inferring} which was collected through annotating 5,000\footnote{Profile counts in the current work differ due to collection limitations (e.g., removed accounts).} English Twitter profiles by crowd-sourcing via Mechanical Turk. This can be considered a `random' sample of Twitter profiles, and is therefore the most unbiased set of the three. Hence, we consider it the most representative of an author profiling set, and employ this as training split (80\%) for $f$, and test split for our attacks (20\%).

The second is the English portion of the Multilingual Hate Speech Fairness corpus of \newcite{huang2020-lrec}, which was collected with a different objective than author profiling. It was aggregated from existing hate speech corpora \cite[by][]{waseem2016hateful,waseem2016you,founta2018large}---which were largely bootstrapped with look-up terms, selection of frequently abusive users, etc.---and annotated post-hoc with demographic information. The collection did not focus on profiles, and most authors are only associated with a single tweet. This can cause a significant domain shift compared to general author profiling. However, it can be seen as freely available (noisy) data.

Lastly, we include a weakly labeled author profiling corpus by \newcite{emmery2017simple}, collected through English keyword look-up for self-reports---similar to \newcite{beller2014ma}. This corpus likely includes incorrect labels, but was collected in less than a day, making it an ideal candidate for realistic access to (new) data to fit the substitute model.

\paragraph{Preprocessing \& Sampling} All three corpora were tokenized using spaCy\footnote{\url{https://spacy.io}} \cite{honnibal2017spacy}. Other than lowercasing, allocating special tokens to user mentions and hashtags (\# and text were split), and URL removal, no additional preprocessing steps were applied. Every author timeline was divided into chunks for a maximum of 100 tweets (i.e., some contain less) to form our documents, implying a maximum of 25 instances per author (some contain one, 2,500 is the API history limit). From the test set, the last\footnote{As the datasets are not shuffled to avoid overfitting on author-specific features, a few documents of the same author might spill from the train into the test split; this avoids incorporating those in our attack sample.} 200 instances were sampled for the attack (110 male, 90 female). While fairly small, this sample does reflect a realistic attack duration and timeline size, as they would be executed for a single profile.

\subsection{Attacks} For the extension of TF, we re-implemented code\footnote{\url{https://github.com/jind11/TextFooler}} by \newcite{jin2019bert} to work with Scikit-learn\footnote{\url{https://scikit-learn.org/}} \cite{pedregosa2011scikit}. For their synonym substitution component, we similarly used counter-fitted embeddings by \newcite{mrkvsic2016counter} trained on Simlex-999 \cite{hill2015simlex}. The USE \cite{cer2018universal} implementation uses TensorFlow\footnote{\url{https://tensorflow.org/}} \cite{abadi2016tensorflow} as back-end, and all BERT-variants were implemented in Hugging Face's\footnote{\url{https://huggingface.co/}} Transformers library \cite{wolf2019transformers} with PyTorch\footnote{\url{https://pytorch.org/}} \cite{NEURIPS2019_9015} as back-end.

We adopt the same parameter settings as \newcite{jin2019bert} throughout our TF experiments: they set $N$ (considered synonyms) and $\delta$ (cosine similarity minimum) empirically to 50 and 0.7 respectively. For MB and DB, we capped $T$ at 50 and top-$k$ at 10 (to improve speed). For DB, we follow \newcite{zhou2019bert} and set the dropout probability to 0.3.

\subsection{Models}

For $f$ and $f'$ we require (preferably fast) pipelines that achieve high accuracy on author profiling tasks, and are sufficiently distinct to gauge how well our attacks transfer across architectures, rather than solely across corpora. As state-of-the-art algorithms have not yet proven to be sufficiently effective for author profiling \cite{joo2019author} we opt for common $n$-gram features and linear models.

\paragraph{Logistic Regression} Logistic Regression (LR) trained on tf$\cdot$idf using uni and bi-gram features proved a strong baseline in author profiling in prior work. The simplicity of this classifier also makes it a substitute model that can realistically be run by an author. No tuning was performed: $C$ is set to $1$.

\paragraph{N-GrAM} The New Groningen Author-profiling Model (N-GrAM) from \newcite{basile2018simply}, was proposed as a highly effective---simple---model that outperforms more complex (neural) alternatives on author profiling with little to no tuning. It uses tf$\cdot$idf-weighted uni and bi-gram token features, character hexa-grams, and sublinearly scaled tf ($1+\log($tf$)$). These features are then passed to a Linear Support Vector Machine \cite{cortes1995support,fan2008liblinear}, where $C=1$.

\subsection{Experimental Setup}

\begin{table}
    \centering
    \footnotesize
    \begin{tabular}{l|p{5cm}}
        \toprule
        data          & Huang, Emmery, Volkova \\
        importance    & Omission score \\
        attack        & Heuristics, TextFooler, Masked BERT, Dropout BERT \\
        model         & Logistic Regression, N-GrAM \\
        ranking       & None, POS + USE, BERT Sim \\
        \bottomrule
    \end{tabular}
    \caption{Grid of possible experimental configurations.}
    \label{tab:exp}
\end{table}

To summarize (and see Table~\ref{tab:exp}), the experiment is conducted as follows: the substitute target model ($f'$)---LR for all experiments---is fit on a given corpus. The real target model ($f$, either LR or N-GrAM) is always fit on the corpus of \newcite{volkova2015inferring}. To evaluate the attacks, a 200-instance sample is used. Target words are ranked via omission scores from $f'$, fed to either our Heuristics, TF, MB, or DB attacks. The heuristics directly change the target words, while the rest outputs a ranked set of replacement candidates. The latter can either be evaluated against $f'$ through the TF pipeline, or the Top-1 candidate is returned. Filtering can be applied through POS/USE for semantic similarity and POS compatibility checks (Check), or not (\st{Check}).

Note that we are predominantly interested in transferability, and would therefore like to test as many combinations of data and architecture access limitations as possible. If we assume an author does not have access to the data, the substitute classifier is trained on any other data than the \citeauthor{volkova2015inferring} corpus. If we assume the author does not know the target model architecture, the target model is N-GrAM (rather than LR). A full model transfer setting (in both data and architecture) will therefore be, e.g.: data $f'$ = \citeauthor{emmery2017simple}, data $f$ = \citeauthor{volkova2014}, $f'$ = LR, and $f$ = NGrAM.  Finally, for comparison to an optimal situation, we test a setting where we do have access to the adversary's data.

\subsection{Evaluation}

\begin{table}
    \centering
    \resizebox{\columnwidth}{!}{
    \begin{tabular}{@{\extracolsep{\fill}}r|r|cc|cc||cc}
        \toprule
        & & \multicolumn{6}{c}{\textbf{test} = \citeauthor{volkova2015inferring} } \\
        \cmidrule(lr){3-8}
        & LR $\ f'$ $\rightarrow$ & \multicolumn{2}{c}{\citeauthor{huang2020-lrec}} &
          \multicolumn{2}{c}{\citeauthor{emmery2017simple}} &
          \multicolumn{2}{c}{\citeauthor{volkova2015inferring}} \\
        \cmidrule(lr){3-4} \cmidrule(lr){5-6} \cmidrule(lr){7-8}
        & $f\ $ $\rightarrow$  & LR                 & NG            & LR              & NG       & LR            & NG  \\
        \midrule   
        
        & none                  &  .885              & .940          & .885           & .940         & .885          & .940    \\
        \cmidrule{1-8}
        \multirow{3}{*}{\rotatebox{90}{Heuristic}}
        & 1337                  &  .770              & .850          & .775           & .835         & .715          & .860    \\
        & flip                  &  .900              & .950          & .885           & .905         & .840          & .905    \\
        & space                 &  .845              & .925          & .760           & .870         & .720          & .850    \\
        \cmidrule{1-8}
        \multirow{3}{*}{\rotatebox{90}{Top-1}}
        & WS                    &  .825              & .930          & .805           & .890         & .750          & .915    \\
        & MB                    &  .655              & .905          & .595           & .785         & .145          & .410    \\
        & DB                    &  .625              & .895          & .575           & .785         & .210          & .530    \\
        \cmidrule{1-8}
        \multirow{3}{*}{\rotatebox{90}{\st{Check}}}
        &  WS                   &  .540              & .855          & .355           & .670          & \textbf{.000}  & \textbf{.009} \\          
        &  MB                   &  \textbf{.415}     & .790          & \textbf{.120}  & \textbf{.420} & \textbf{.000}  & .085    \\          
        &  DB                   &  .430              & \textbf{.775} & .175           & .430          & \textbf{.000}  & .085    \\          
        \cmidrule{1-8}
        \multirow{3}{*}{\rotatebox{90}{Check}}
        & TF                    &  .705              & .920           & .780          & .910          & .375          & .700    \\
        & TF + MB               &  .640              & .880           & .760          & .890          & .380          & .725    \\          
        & TF + DB               &  .650              & .885           & .755          & .890          & .435          & .715    \\          
        \bottomrule
    \end{tabular}}
    \caption{Post-attack accuracy scores (below chance (55\%) = better) of $f$ on a test sample from the \citeauthor{volkova2015inferring} corpus. Left, the attack conditions: heuristics, top-1 synonym, applying POS and USE similarity checks, or not applying those checks (\st{Check}). Splits per training corpus are noted for $f^\prime$ (always Logistic Regression (LR)). As target model, either LR, or N-GrAM (NG) was used. The substitution attacks are TextFooler (TF), Masked (MB) and Dropout BERT (DB). If TF's stopping criterion was used, TF + is noted. Word Similarity (WS), reflects the TF pipeline without checks.}
    \label{tab:results}
\end{table}

\paragraph{Metrics} The obfuscation success is measured as any accuracy score below chance level performance, which given our test sample is 55\%. We would argue that random performance is preferred in scenarios where the prediction of the opposite label is undesired. For the current task, however, any accuracy drop to around or lower than chance level satisfies the conditions for successful obfuscation.\footnote{If an attack drops accuracy to 0\%, this effectively flips (in case of a binary label) the label. This label might \emph{also} be undesired by the author (e.g., being classified as having polar opposite political views). This implies the target model being maximally unsure about the classification is desirable.} To evaluate the semantic preservation of the attacked sentences, we calculate both \textsc{meteor} \cite{banerjee2005,lavie2009meteor} using \texttt{nltk},\footnote{\url{https://www.nltk.org/_modules/nltk/translate/meteor_score.html}} and BERTScore \cite{zhang2019bertscore} between $D$ and $D_\textsc{adv}$. \textsc{meteor} captures flexible uni-gram token overlap including morphological variants, and BERTScore calculates similarities with respect to the sentence context.

\begin{table}
    \centering
    \small
    \begin{tabular}{l|l|rr} % ||rrrrrrrrrrrr}
    \toprule
    & \citeauthor{volkova2015inferring} $\rightarrow$  &    \textsc{train} & \textsc{test} \\
    \midrule
    \multirow{2}{*}{\rotatebox{90}{\textsc{train}}}
    & \citeauthor{huang2020-lrec}         &  0.640   & 0.620         \\ [0.2cm]
    & \citeauthor{emmery2017simple}       &  0.725   & 0.890         \\
    \bottomrule
    \end{tabular} % }
    \caption{Gender prediction accuracies of the substitute models $f^\prime$ on train and test splits of $f$.} 
    \label{tab:dom}
\end{table}

\paragraph{Human Evaluation} For the human evaluation, we sampled 20 document pieces (one or more tweets) for each attack type in the best performing experimental configuration. A piece was chosen if it satisfied these criteria: i) contains changes for all three attacks, ii) consists of at least 15 words (excluding emojis and tags), and iii) does not contain obvious profanity.\footnote{To avoid exposing the raters to overly toxic content, blatant examples were filtered using a keyword list. Some minor examples remained, for which we added a disclaimer.} All 60 document pieces of the three models were shuffled, and the 20 original versions were appended at the end (so that `correct' pieces were seen last). Each substitute model therefore has 80 items for evaluation.

While in prior work it is common to rate semantic consistency, fluency, and label a text \cite[see e.g.,][]{potthast2016,jin2019bert}, our Twitter data are too noisy (including many spelling and grammar errors in the originals), and document batches too long to make this a feasible task. Instead, our six participants (three per substitute) were asked to indicate if: a) a sentence was artificially changed, and if so, b) indicate one word that raised their suspicion. This way, we can evaluate which attack produces the most natural sentences, and the least obvious changes to the input. 

The items were rated individually; the human evaluators did not know beforehand that different versions of the same sentences were repeated, nor that the originals were shown at the end. All participants have a university-level education, a high English proficiency, and are familiar with the domain of the data. Several example ratings of the same sentence can be found in Table~\ref{tab:example}.

\section{Results}

\subsection{Domain Shift} As we alluded to in Section~\ref{subs:data}, both corpora used to train our substitute models were in fact not reference corpora for author profiling, and can therefore be considered as suboptimal, disjoint domains. The \citeauthor{huang2020-lrec} corpus in particular shows a strong domain shift (see Table~\ref{tab:dom}) for both training and test sets. The distantly labeled \citeauthor{emmery2017simple} corpus achieves 7.5\% more accuracy on the train split of \citeauthor{volkova2015inferring}, and test performance is significantly higher (27\%). We might therefore expect better obfuscation performance from the latter.

\subsection{Baselines} The results for all attacks are shown in Table~\ref{tab:results}. Note that these are performances for $f$; therefore, when no attacks are applied (none), the performance for both substitute corpora stays the same (as those only influence the attacks). For the heuristic attacks, 1337 seems to make the more robust baseline; outperforming some of the other settings---even on transferability. A surface-level advantage is that this attack has a minor impact on readability (when applied conservatively) and does not change semantics; however, the heuristic attacks are fairly simple to mitigate in preprocessing \cite{juola2011analyzing} and through character features (as shown by the performance of the N-GrAM model). For transferability, we evidently need to do more than simply trying to convert words to be out-of-vocabulary (OOV) with noise. While it can be argued the heuristics could change all words, shifting everything OOV would not be robust; the target model side could easily spot the anomalous input and might act (e.g., reject) accordingly.

\begin{figure}
    \centering
    \begin{minipage}{.33\columnwidth}
        \includegraphics[width=6.5em, clip]{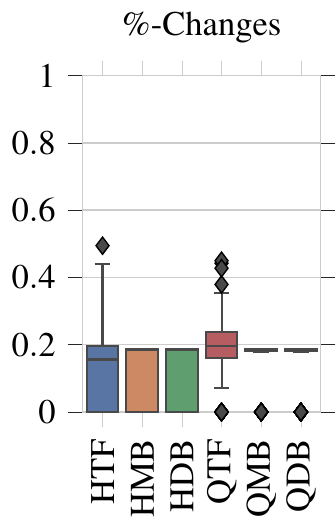}
    \end{minipage}
    \begin{minipage}{.30\columnwidth}
        \includegraphics[width=6.5em, clip]{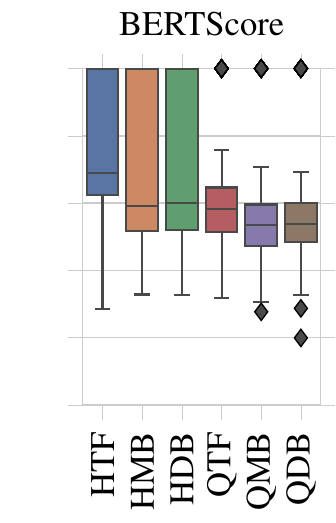}
    \end{minipage}
    \begin{minipage}{.30\columnwidth}
        \includegraphics[width=6.5em, clip]{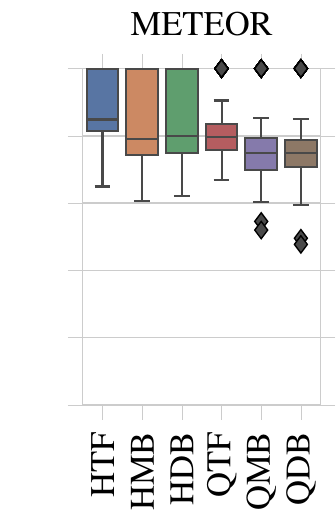}
    \end{minipage}
      \captionof{figure}{Rewrite metrics for the three different attacks: TextFooler (TF), Masked (MB) and Dropout BERT (DB)---split by corpora \citeauthor{huang2020-lrec} (H), \citeauthor{emmery2017simple} (Q). Shown are the relative number of changes (by document length), BERTScore ($F_1$), and \textsc{meteor} (MET) with respect to the original document. 
      }
    \label{fig:eval}
\end{figure}

\subsection{Attack Transferability} Transferability can be assessed by comparing the LR and N-GrAM (NG) columns. Globally it can be observed that the substitute models trained on the \citeauthor{emmery2017simple} corpus systematically outperform those trained on \citeauthor{huang2020-lrec}; both for the settings where the adversary's architecture is known (LR), and where it is unknown (NG). This matches our expectations from the observed domain shift. Our results also show that a noticeable decrease in obfuscation performance occurs (10-30\% increased target model performance) when the attacks are transferred to different data and another model. In contrast, as can be observed from the last two columns in Table~\ref{tab:results}, in a practically unrealistic setting where the model and data are available, the obfuscation is specifically tailored to known weaknesses and therefore highly effective.\footnote{Jin et al. (2020) found similar drops to 0\% accuracy with a comparable percentage of changes for word-level models.}

\subsection{Transformer Performance} Looking at the Top-1, \st{Check} and Check brackets (Table~\ref{tab:results}), other than the BERT-based models having higher success of transferability than TF, they also retain obfuscation success; deteriorating the target model's performance to lower than chance level (55\%) for the settings not using additional checks. This comparison also demonstrates the synonym ranking to work (Top-1 vs. \st{Check} and Check), and the Check condition to be too restrictive; attaining lower attack power, and low transferability. This is further illustrated by the \%-changes shown in Figure~\ref{fig:eval}. Comparing the MB and DB variants, their performance seems almost identical, with masking having a slight advantage. As \newcite{zhou2019bert} argued, applying dropout should produce words that are closer to the original (compared to MB), which might affect obfuscation performance. Additionally, the BERT similarity ranking (described in Section~\ref{p:bertsim}) applied to the Masked substitution candidates could have some beneficial effect. This will have to be studied in more detail using the output evaluations.

\paragraph{Rewrite Metrics} 

\begin{table}
    \centering
    \resizebox{\columnwidth}{!}{
    \begin{tabular}{l|r|rrr|rrr}
    \toprule
    & & \multicolumn{3}{c}{\citeauthor{huang2020-lrec}} & \multicolumn{3}{c}{\citeauthor{emmery2017simple}} \\
    \cmidrule(lr){3-5} \cmidrule(lr){6-8}
                     &  ORG & TF   & MB   & DB   & TF   & MB   & DB  \\
    \midrule
    \textsc{altered} & .888 & .967 & .633 & .783 & .950 & \textbf{.617} & .633  \\
    \textsc{word}    &   -  & .950 & .583 & .700 & .867 & \textbf{.433} & \textbf{.433}  \\
    \bottomrule
    \end{tabular}}
    \caption{Human accuracy scores of predicting if a text was altered, and guessing the attacked word (lower is better). All substitute models are those with the \st{Check} setting, trained on different corpora (i.e., different words are attacked per training corpus). ORG indicates correct prediction of the originals.}
    \label{tab:huval}
\end{table}

\begin{table*}
\def\arraystretch{1.4}
\small
\resizebox{\textwidth}{!}{
\begin{tabular}{p{0.8cm}|p{15cm}}
\textsc{org} & ready to go home already . a better relationship with god \includegraphics[width=0.7em]{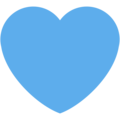} \includegraphics[width=0.7em]{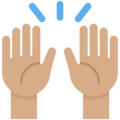} . i need another job asap . \\
\textsc{htf} & \textbf{loan} to go \textbf{houses} already . a \textbf{improved} \textbf{relations} with \textbf{jesus} \includegraphics[width=0.7em]{emoji-heart.png} \includegraphics[width=0.7em]{emoji-hands.png} . i \textbf{should} another \textbf{labour} asap . \\
\textsc{hmb} & ready to go \textbf{on} already . a better relationship with god \includegraphics[width=0.7em]{emoji-heart.png} \includegraphics[width=0.7em]{emoji-hands.png} . i need another \textbf{guy} \textbf{man} . \\
\textsc{hdb} & ready to go \textbf{somewhere} already . a better relationship with god \includegraphics[width=0.7em]{emoji-heart.png} \includegraphics[width=0.7em]{emoji-hands.png} . i need another \textbf{position} \textbf{vs} . \\ [0.2cm]
\textsc{org} & trump criticizes kim jong un after missile launch : ‘ does this guy have anything better to do ? ’ . \\
\textsc{htf} & \textbf{tramp criticized kam yung jt} after \textbf{rocket start} : ‘ does this \textbf{boyfriend} have anything \textbf{best} to do ? ’ . \\
\textsc{hmb} & trump criticizes \textbf{ha woman congressman} after \textbf{campaign} launch : ‘ does this \textbf{book} have anything \textbf{else} to do ? ’ . \\
\textsc{hdb} & trump criticizes \textbf{in} \textbf{at} \textbf{sin} after \textbf{bomb} launch : ‘ does this \textbf{kid} have anything \textbf{less} to do ? ’ . \\

\end{tabular}}
\caption{Example ratings of different attacks (not shown together to the human evaluators) on two sentences with varying semantic consitency and human detection accuracy. In the first example, \textsc{hmb} was marked unaltered by all raters, \textsc{hdb} by the majority, and \textsc{htf} by none. In the second, only \textsc{hdb} was marked unaltered, by only one rater. Attacked words are marked in bold, guessing any one of these would count as correctly identifying the attack.}
\label{tab:example}
\end{table*}

The metrics in Figure~\ref{fig:eval} show a common initial limitation in their application to this task: the more frequent an attack makes no changes, the higher the automatic evaluation metrics (BERTScore, \textsc{meteor}). Hence, to compare models, these scores need to be considered in light of the obfuscation performance, and related work. It can be observed that with consistently higher changes, MB and DB score lower on semantic consistency than TF. However, between MB and DB, and TF for the \citeauthor{emmery2017simple} corpus, these differences are minor. Furthermore, despite being fit on a different domain, these scores are comparable to prior obfuscation work (e.g., \newcite{shetty2018a4nt} show \textsc{meteor} scores between 0.69 and 0.79).

\paragraph{Human Evaluation}

The results in Table~\ref{tab:huval} reflect the same trend that can be observed in Table~\ref{tab:results}; high obfuscation success seems to result in higher human error when predicting if a sentence was obfuscated. Conversely, it seems that despite higher semantic consistency scores, the original TF pipeline is easier to detect. This can be attributed to the number of spelling and grammar errors the model makes without its additional checks. Furthermore, the 11\% error in identifying the original sentences also reflects some expected margin of error in this task, as our Twitter data is inherently noisy. Finally, while these results are in line with the obfuscation success, and are lower than detectability scores in related work \cite{mahmood2020girl}, they also indicate that the models are still detectable above chance-level. Given three alternatives (including the original), performance should be 25\% or lower to indicate no intrusive changes are made to text (that are not semantically coherent or not inconspicuous enough---both metrics used by \citeauthor{potthast2016}, \citeyear{potthast2016}). Therefore, while the presented approaches are effective, and realistically transferable, there is room for improvement for practical applicability.

\section{Discussion and Future Work}

We have demonstrated the performance of author attribute obfuscation under a realistic setting. Using a simple Logistic Regression model for candidate suggestion, trained on a weakly labeled corpus collected in a day, the attacks successfully transferred to different data and architectures. This is a promising result for future adversarial work on this task, and its practical implementation.

It remains challenging to automatically evaluate how invasive the required number of changes are for successful obfuscation---particularly to an author's message consistency as a whole. However, in practice such considerations could be left up to the author. In this human-in-the-loop scenario, a more extensive set of candidates could be suggested, and their effect on the substitute model shown interactively. This way, the attacks can be manually tuned to find a balance of effectiveness, inconspicuousness, and to guarantee semantic consistency. It would also show the author how their writing style affects potential future inferences.

Regarding the performance of the attacks: we demonstrated the general effectiveness of contextual language models in retrieving candidate suggestions. However, the quality of those candidates might be improved with more extensive rule-based checks; e.g., through deeper analyses using parsing. Nevertheless, such venues leave us with a core limitation of rewriting language, and therefore more broadly NLP: while the Masked attacks seemed more successful in our experiments, after manual inspection of the perturbations Dropout was found to often be semantically closer (see also Table~\ref{tab:example})---which was not reflected in the human evaluation. This begs the question if \emph{any} automated approach, evaluated under the current limitations of semantic consistency metrics, could realistically optimize for both obfuscation and inconspicuousness.

As such, we would argue that future work should focus on making as few perturbations as possible, retaining only the minimum amount of required obfuscation success. Given this, the other constraints become less relevant; one could generate short sentences (e.g., a single tweet) that might be semantically or contextually incorrect, but if it is a message in a long post history, it will hardly be detectable or intrusive. This would require certain triggers (as demonstrated by \newcite{wallace2019universal} for example), and ascertaining how well they transfer. 

\section{Conclusion}

In our work, we argued realistic adversarial stylometry should be tested on transferability in settings where there is no access to the target model's data or architecture. We extended previous adversarial text classification work with two transformer-based models, and studied their obfuscation success in such a setting. We showed them to reliably drop target model performance below chance, though human detectability of the attacks remained above chance. Future work could focus on further minimizing this detection under our realistic constraints.

\section*{Acknowledgments}

Our research strongly relied on openly available resources.  We thank all whose work we could use. We would also like to thank the anonymous reviewers, Bertrand Higy, Bram Willemsen, and Chris van der Lee for their valuable feedback. Ákos Kádár contributed to this research independently. This work does not reflect Borealis AI’s views nor any information Ákos Kádár may have learned while employed by Borealis AI.

\bibliographystyle{acl_natbib}
\bibliography{refs}

\end{document}